\pdfoutput=1

\documentclass[11pt]{article}

\usepackage{acl}

\usepackage{times}
\usepackage{latexsym}
\usepackage{array}

\usepackage[T1]{fontenc}

\usepackage[utf8]{inputenc}
\usepackage{graphicx}
\usepackage{booktabs}
\usepackage[inline]{enumitem}
\usepackage{makecell}
\usepackage{header}
\usepackage{tikz}
\usetikzlibrary{shapes.geometric, arrows}
\usepackage{makecell}

\usepackage{tabularx}
\usepackage{array}
\usepackage{microtype}
\usepackage{pifont}
\newcommand{\cmark}{\ding{51}}%
\newcommand{\xmark}{\ding{55}}%
\newcolumntype{C}[1]{>{\centering\arraybackslash}p{#1}}
\usepackage{arydshln}
\usepackage{graphicx}
\usepackage{booktabs}
\usepackage{multirow}
\usepackage{multicol}
\usepackage{microtype}
\usepackage{array}
\usepackage{inconsolata}
\usepackage{adjustbox}
\usepackage{soul}
\usepackage{amsmath}
\usepackage{hyperref}
\usepackage{changepage}
\usepackage{arydshln}
\usepackage{pifont}

\title{\textit{What if you said that differently?}: How Explanation Formats Affect\\Human Feedback Efficacy and User Perception}

\author{
    Chaitanya Malaviya,
    Subin Lee,
    Dan Roth,
    Mark Yatskar \\
    University of Pennsylvania\\
    {\tt \{cmalaviy,subinlee,danroth,myatskar\}@upenn.edu}
}

\begin{document}
\maketitle

\begin{abstract}

Eliciting feedback from end users of NLP models can be beneficial for improving models. However, \textit{how should we present model responses to users so they are most amenable to be corrected from user feedback}? Further, what properties do users value to understand and trust responses? We answer these questions by analyzing the effect of rationales (or explanations) generated by QA models to support their answers.
 
We specifically consider decomposed QA models that first extract an intermediate rationale based on a context and a question and then use solely this rationale to answer the question.
A rationale outlines the approach followed by the model to answer the question.
Our work considers various formats of these rationales that vary according to well-defined properties of interest.
We sample rationales from language models using few-shot prompting for two datasets, and then perform two user studies.
First, we present users with incorrect answers and corresponding rationales in various formats and ask them to provide natural language feedback to revise the rationale. 
We then measure the effectiveness of this feedback in patching these rationales through in-context learning.
The second study evaluates how well different rationale formats enable users to understand and trust model answers, when they are correct.
We find that rationale formats significantly affect how easy it is (1) for users to give feedback for rationales, and (2) for models to subsequently execute this feedback. 
In addition, formats with attributions to the context and in-depth reasoning significantly enhance user-reported understanding and trust of model outputs.\footnote{Code and data is available at \url{https://github.com/chaitanyamalaviya/rationale_formats}.}

\end{abstract}

\section{Introduction}


\begin{figure}[t!]
    \centering
    \includegraphics[width=\columnwidth,height=7cm,keepaspectratio]{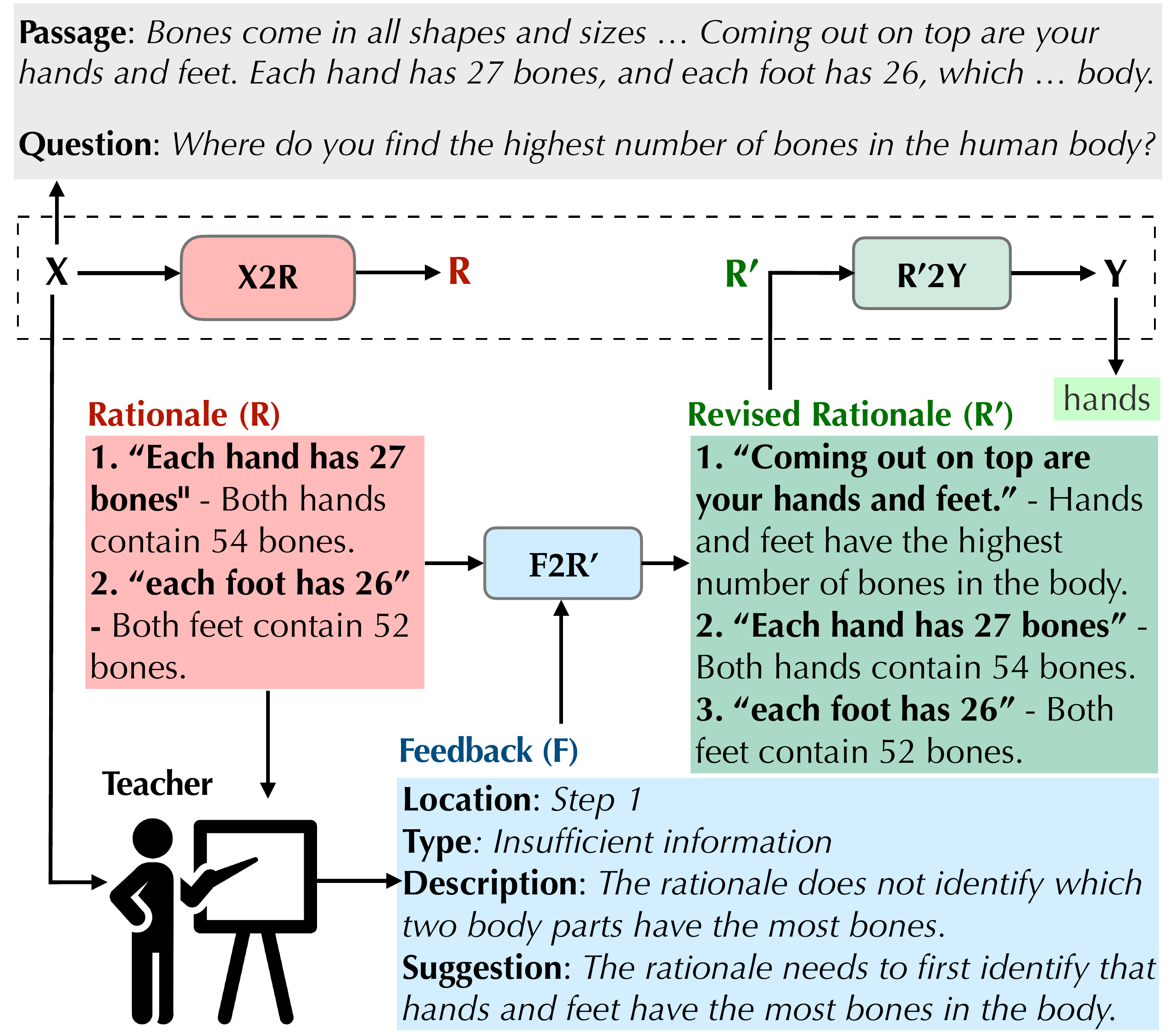}
    \caption{The framework for incorporating human feedback into decomposed QA models. A model $X2R$ generates a rationale $R$ to answer a question based on a passage. A human teacher then provides natural language feedback for $R$, which is used to generate a revised rationale $R'$ from $F2R'$.
    Finally, this revised rationale is used to generate the final answer $Y$. We study various formats of the intermediate rationale $R$.}
    \vspace{-10pt}
    \label{fig:workflow}
\end{figure}

Question answering models can often be black boxes, as their reasoning process is mostly opaque to model builders as well as end users.
This can inhibit the ability of users to provide helpful critiques to models to repair them.
Generating \textit{rationales} (or explanations) along with answers is a viable approach that can alleviate these concerns, but these rationales are inherently not faithful and can sometimes be inconsistent with the answers themselves \cite{ye2022unreliability,turpin2023language,lanham2023measuring,radhakrishnan2023question}.

This motivates approaches that decompose the question answering task into two stages (depicted with dashed lines in Figure~\ref{fig:workflow}), where we first generate a rationale for the question using the given context ($X2R$), then use only this rationale to answer the question ($R2Y$) \cite{lei-etal-2016-rationalizing,eisenstein2022honest}. A rationale may provide a justification for the answer by presenting an outline for how the question can be answered. By only relying on the rationale as context, the answer generation model ($R2Y$) has a stronger inductive bias to generate an answer that is faithful to the rationale. 

Crucially, faithful rationales can allow users to follow the model's line of reasoning, and subsequently provide actionable feedback to the model. This feedback can be used to repair individual outputs or enable generalization to novel instances.
However, it is unclear precisely how a rationale should be formatted, i) to best aid the user's understanding of the model's reasoning, and ii) their ability to provide feedback for the response. 

Our work specifically addresses the question of how we can format intermediate rationales ($R$) for decomposed QA systems, such that they are \textit{easy to repair through human feedback}. Further, we analyze what properties make rationales \textit{interpretable}, and \textit{trustworthy} to users.
Previous work on decomposed question answering mostly consider rationales as text snippets extracted from the context, optionally marked up with coreferences that make the snippets standalone \cite{deyoung-etal-2020-eraser,eisenstein2022honest}.
Although extractive snippets can serve useful for providing minimal context that rationalizes an answer, they do not provide much insight into the model's reasoning process. 
This may limit a user's understanding and their ability to critique the model.
We consider alternative formats of rationales, which vary according to well-defined characteristics (\S\ref{sec:rationales}). Examples of these formats and how they vary are presented in Table~\ref{tab:formats}.

Based on the considered rationale formats, we generate responses (rationales \& answers) from a decomposed QA system.
We then perform two user studies where we measure i) the effectiveness of user feedback in patching rationales in different formats and ii) the ability of different formats to enable users to understand and trust responses.
In our first study, we sample \textit{incorrect} answers corresponding to all rationale formats, and ask annotators to provide natural language feedback for the rationales (\S\ref{sec:study1}).
We use this feedback to then revise the rationales ($F2R'$) and regenerate the final answer ($R'2Y$). Comparing the regeneration accuracy with different rationale formats provides insight into properties of rationales that make them easy to repair. Further, they provide an upper bound for how much improvement can be expected through automated feedback by repairing rationales \cite{chen2023teaching,madaan2023self}. In our second study, we elicit judgements of interpretability and trustworthiness for \textit{correct} answers and their accompanying rationales from users (\S\ref{sec:study2}).

We consider two tasks: general reading comprehension (Quoref; \citet{dasigi-etal-2019-quoref}) and medical reading comprehension (PubMedQA; \citet{jin-etal-2019-pubmedqa}). Each block in our overall workflow (Figure~\ref{fig:workflow}) is implemented through few-shot prompting of a large language model. Our experiments suggest that rationale formats significantly affect i) users' ease of providing feedback and ii) the model's ability to execute that feedback. In addition to being critiquable, certain rationale formats are more helpful in aiding users to understand and trust the model's answers. One such format is the \textit{annotated report}, which includes a list of extractive phrases and free-text inferences based on each phrase. Finally, among a few characteristics of rationales presented to users, we find that users rate attribution and depth of reasoning as characteristics that are most important to them.


\begin{figure*}[t!]
    \centering
    \includegraphics[width=2\columnwidth,height=7cm,keepaspectratio]{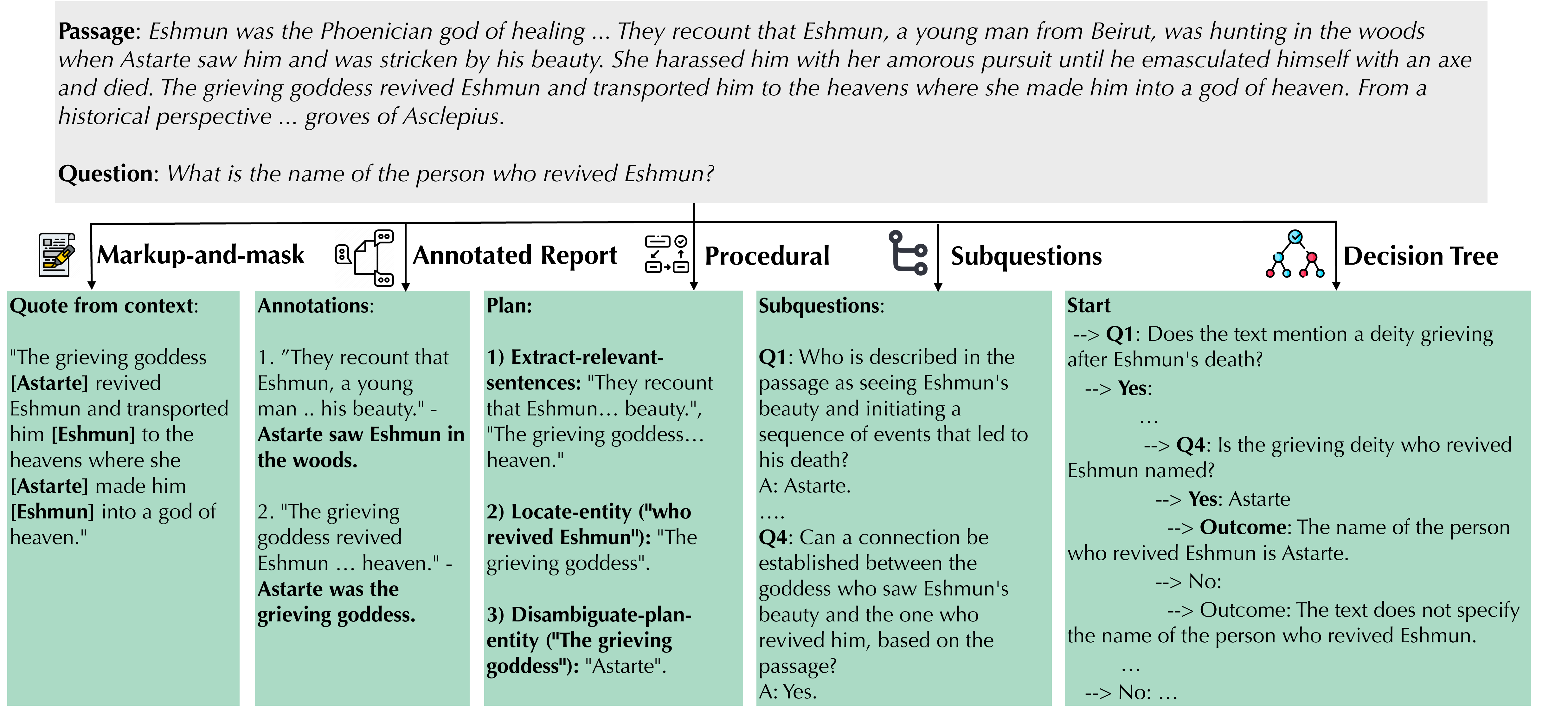}
    \caption{Examples of the different rationale formats considered for representing intermediate rationales.}
    \label{fig:formats}
\end{figure*}

\section{Problem Formulation}


Consider a standard reading comprehension task, where we are given a passage, a question based on this passage, and a reference answer $Y$ as a labeled example in our dataset.
We represent the input information in the task (the passage and the question) with $X$.
Typically, we would train a model $X2Y$ that predicts $Y$ given $X$ by learning $P(Y|X)$.
Assuming $X2Y$ is a black-box model, without loss of generality, the model may internally compute a latent representation, which is usually not extractable in an interpretable format, from $X$ to predict $Y$, internally decomposing the problem. 
This can restrict the transparency of the model because we cannot provide a faithful reasoning to an end user that supports the answer.

\subsection{Decomposed QA Pipeline}

In the decomposed QA pipeline, we factor the QA problem in the following manner (see Figure~\ref{fig:workflow} for an illustration):
\begin{equation}
P(Y,R|X) = P(Y|R,X) P(R|X)
\end{equation}

Since Y is independent of X given R (assume Q is part of R for simplicity), we have,
\begin{equation}
P(Y,R|X) = P(Y|R) P(R|X)
\end{equation}

Let’s first consider a rationale-generating model $X2R$, that generates a textual rationale $R$ given $X$. This rationale provides an outline of the approach proposed by the model to answer the question. Consider also an answer-generating model $R2Y$ that generates an answer ${\hat{Y}}$ given the predicted rationale $R$. In this decomposed model, $R2Y$ has a strong inductive bias to use the information presented in $R$ for its reasoning. 
Further, $R$ can be explicitly shown to an end user, which increases the transparency of the entire system.

\subsection{Debugging the Decomposed QA Pipeline}

Next, let’s assume a set of spans $X_s$ from $X$ that are sufficient to predict the answer $Y$. First, we note that for the answer to be correct, i.e. $\hat{Y}=Y$, the predicted rationale $R$ must contain all the information contained in $X_s$, i.e., $I(X_s)$.
Errors in answers generated by $R2Y$ can be a result of (1) insufficient or incorrect context, when $I(X_s) \not\subseteq I(R)$, and / or (2) limited model capacity of $R2Y$, when $I(X_s) \subseteq I(R)$ .
Repairing the modeling pipeline (i.e., $X2R$ $+$ $R2Y$) can hence either involve improving the quality of generated rationales $R$ produced by $X2R$ or improving the modeling capacity of $R2Y$. 
Note that the example may require information beyond what is in the passage (for example, domain knowledge or commonsense knowledge).

We consider the scenario where we repair the rationales generated by $X2R$ through natural language feedback. We assume a teacher $T$ who writes feedback for generated rationales $R$, where they describe flaws in $R$. Generated rationales can be lacking in various ways: 1) insufficient information: $R$ may not contain crucial information required to perform the inference (i.e., $I(X_s) \not\subseteq I(R)$), or 2) incorrect information: $R$ may contain hallucinated content or incorrect reasoning chains that could mislead the answer-generating model.

The teacher $T$ in our case is an end user, who could optionally be a domain expert depending on the task. We evaluate whether the format of predicted rationales $R$ is interpretable and easy to repair for $T$. Based on $T$'s feedback $F_k$ for a subset of examples $X_k \subset X$, we revise the initial rationale $R$ to generate a revised rationale $R’$ using a model $F2R'$. This revised rationale is then used by $R'2Y$ to generate the final answer.

\section{Intermediate Rationale Formats}
\label{sec:rationales}

\begin{table*}[!ht]
\centering
\footnotesize
\scalebox{.85}{
\rowcolors{2}{blue!5}{blue!5}
\begin{tabular}{ C{3cm}|p{5cm}|C{2cm}|C{2cm}|C{2cm}|C{2cm} }
\textbf{Format} & \textbf{Description} & \textbf{Attribution Provided} & \textbf{Reasoning Exposed} & \textbf{Sequential Reasoning} & \textbf{Free-text annotations} \\ \hline
\textbf{Markup-and-Mask} & Quoted sentences from the context, marked up with coreferences for pronouns and ambiguous phrases & \cmark & \xmark & \xmark & \xmark  \\
 \textbf{Annotated Report} & Quoted phrases from the context and an inference from each phrase & \cmark & \cmark & \cmark & \cmark  \\
 \textbf{Procedural} & Step-by-step plan for solving the question with pre-defined operations & \cmark & \cmark & \cmark & \xmark  \\
 \textbf{Subquestions} & Breakdown of the original question into subquestions & \xmark & \cmark & \cmark & \cmark \\
 \textbf{Decision Tree} & Breakdown of the original question into subquestions presented in a tree structure, with Yes/No outcomes for each subquestion & \xmark & \cmark & \cmark & \cmark \\
 \hline
\end{tabular}
}
\caption{Descriptions of the rationale formats considered in our work and the characteristics along which they differ.} 
\label{tab:formats}
\end{table*}

Rationales in NLP tasks are usually presented as compressed text snippets extracted from the given input \cite{deyoung-etal-2020-eraser,eisenstein2022honest}. However, text snippets from the context alone may not make the model’s reasoning explicit and transparent to users. We consider alternative rationale formats that describe the model's reasoning. We describe these formats below. A summary of these formats and how they vary according to rationale properties, is given in Table~\ref{tab:formats}.

\subsection{Rationale Formats}
\paragraph{Markup-and-Mask (\texttt{markup\_mask}).} This format, proposed by \citet{eisenstein2022honest}, extracts sentences from the context that are relevant to answering the question. Sentences are decontextualized by markups that resolve coreferences and other ambiguous phrases \cite{choi-etal-2021-decontextualization}.

\paragraph{Annotated Report (\texttt{annotated\_report}).} The annotated report extracts phrases from the context and generates a free-text inference based on each phrase that is relevant to answering the question. This is broadly inspired by the way readers highlight and annotate key spans in documents while reading (also found as marginalia in books).

\paragraph{Procedural (\texttt{procedural}).} A procedural rationale is a step-by-step plan that uses predefined operations to answer the question. Similar works that broadly propose a plan-based rationale have been explored in prior work, in different contexts \cite{sun2023pearl,wang-etal-2023-plan}. The primitive operations we consider include an operation to extract relevant sentences, disambiguate an entity from the question or the plan so far, and locate an entity by answering a subquestion. These are further described in Appendix~\ref{app:exps}.

\paragraph{Subquestions (\texttt{subquestions}).} Subquestions simply decompose the original question into multiple questions that provide relevant information to answer the question. These have been explored as a form of rationale in various works \cite{geva2021did,khot-etal-2021-text,press2022measuring,dua-etal-2022-successive,zhou2022least}.

\paragraph{Decision Tree (\texttt{decision\_tree}).} We also consider a tree-structured rationale, inspired by fast-and-frugal trees \cite{martignon2003naive} as well as prompting work that explores tree-like structures \cite{yao2023tree}. This format decomposes the original question into Yes/No subquestions in a tree-like structure and also shows the incorrect tree traversals for completeness.

\section{Experimental Setup}

\subsection{Datasets}

We consider two datasets for our studies: Quoref \cite{dasigi-etal-2019-quoref} and PubMedQA \cite{jin-etal-2019-pubmedqa}. The first is a general reading comprehension dataset while the second involves medical reading comprehension. In contrast to Quoref, PubMedQA often requires domain-specific knowledge for answering the question. We use all validation set examples of Quoref (2418 examples) and all labeled examples in PubMedQA (1000 examples)

\subsection{Sampling Rationales and Answers}

We sample rationales and answers for all 5 formats in a decomposed QA pipeline, where both $X2R$ and $R2Y$ are implemented using few-shot prompting. We first prompt \texttt{gpt-3.5-turbo} for rationales by providing the passage and question. The question and only the generated rationales are then used to prompt the same model to generate the final answer. In both cases, we sample few-shot exemplars using BM25 \cite{robertson2009probabilistic} from a set of 100 manually labeled examples with rationales. We sample as many exemplars as can fit within the maximum sequence length (4096) of the model. This usually amounts to 3-5 exemplars. The prompts used and other hyperparameters are provided in Appendix~\ref{app:exps}.




\section{Study 1: Repairing Rationales through Human Feedback}
\label{sec:study1}

\subsection{Setup}
In this study, we measure the critiquability or ease of repair of different rationale formats. This is done by collecting natural language feedback from human annotators for rationales corresponding to incorrect answers. We sample examples for which the decomposed QA pipeline predicts \textit{incorrect answers} for all 5 rationale formats. 
In all, we collect 490 feedback statements for Quoref and 555 feedback statements for PubMedQA.\footnote{This corresponds to 5 rationale formats and 98 examples for Quoref and 111 examples for PubMedQA.}

In each annotation task, annotators are shown a single example (question \& passage) with all 5 rationale formats and their corresponding answers. This controls for annotator variance and variance across examples. To control for ordering effects, we randomize the order in which the rationale formats are presented to annotators. For each rationale format, annotators are asked to write natural language feedback to repair the rationale.

We use this natural language feedback to generate the revised rationale $R'$. To do this, we prompt \texttt{gpt-3.5-turbo} with the passage, the question, the original rationale, and human-written feedback. Finally, we generate the final answer by few-shot prompting the same model using just the revised rationale and question. Prompts and other hyperparameters are in Appendix~\ref{app:exps}.

\subsection{Participants}

For this study as well as the study in section~\ref{sec:study2}, we recruit participants through Prolific. Participants are required to be fluent in English and are based primarily in English-speaking countries. For PubMedQA examples, they are required to be working in the healthcare sector. 
For more details, please see Appendix~\ref{app:annotation}.

\subsection{Task}

To prime annotators for formulating their feedback, we ask them to first evaluate the sufficiency and faithfulness of each rationale to the context. They label these two properties for each rationale format on a Likert scale (the precise descriptions of the options in all Likert-scale questions are in Figure~\ref{fig:interface2}).

\paragraph{Sufficiency.} Annotators evaluate if the rationale provides enough information to answer the question, without the context. Note that the rationale may contain inaccuracies but still be sufficient. 
They mark sufficiency as (\textit{Sufficient}, \textit{A bit insufficient}, \textit{Entirely insufficient}).

\paragraph{Faithfulness to context.} Next, annotators evaluate whether the rationale accurately draws conclusions from the context without misrepresenting any information. They mark faithfulness on a scale of (\textit{Accurate}, \textit{A bit inaccurate}, \textit{Very inaccurate}).

\subsubsection{Feedback}

Annotators are asked to formulate natural language feedback that would be most useful in directing the model to the reference correct answer. It could target missing or incorrect information in the rationale, but cannot explicitly reveal the correct answer. Feedback is elicited in multiple steps (examples of feedback written by annotators are in Table~\ref{tab:feedback_examples_quoref}, \ref{tab:feedback_examples_pubmedqa}):


\begin{enumerate}
    \setlength{\itemsep}{0pt}
    \item \textbf{Location of error}: Annotators are required to list the step(s) (or question number) in which the error occurs.
    \item \textbf{Type of error}: Annotators then identify the type of the error. We show them a few common error types that occur in rationales (for example, insufficient information, irrelevant information, incorrect inferences etc).
    \item \textbf{Description of error}: Next, annotators use concrete details from the rationale, question \& context to provide a description of the error.
    \item \textbf{Actionable suggestion}: Finally, annotators provide an actionable edit that would repair the rationale, again using concrete details from the rationale, question \& context.
\end{enumerate}

\paragraph{Ease of repair.} Using annotator feedback, we can measure how amenable each rationale format is for repair. However, this does not reflect annotators' ease of providing feedback for each format. 
We elicit this directly on a scale of (\textit{Very easy}, \textit{Somewhat easy}, \textit{Somewhat hard}, and \textit{Very hard}).

\begin{table}[!t]
\centering
\scalebox{0.85}{
\begin{tabular}{lccc}
    \toprule
    & \multicolumn{2}{c}{Quoref} & PubMedQA \\
    \cmidrule(lr){2-3} \cmidrule(lr){4-4}
    \textbf{Rationale Format} & \textbf{EM} & \textbf{F1} & \textbf{Accuracy} \\ 
    \midrule
    \texttt{none} & 70.31 & 79.65 & 69.30 \\ \hline
    \texttt{markup\_mask} & 57.44 & 68.10 & 62.20 \\
    \texttt{annotated\_report} & 60.26 & 70.20 & 70.20 \\
    \texttt{procedural} & 66.09 & 77.05 & 68.30 \\
    \texttt{subquestions} & 54.26 & 63.05 & 68.90 \\
    \texttt{decision\_tree} & 68.61 & 77.09 & 46.70 \\
    \bottomrule
\end{tabular}}
\caption{Initial scores using the decomposed QA pipeline ($X2R$ + $R2Y$) for different rationale formats.}
\label{tab:initial_res}
\end{table}

\begin{table*}[!ht]
\centering
\small
\begin{tabular}{lcccccc}
    \toprule
    & \multicolumn{3}{c}{Quoref} & \multicolumn{3}{c}{PubMedQA} \\
    \cmidrule(lr){2-4} \cmidrule(lr){5-7}
    \textbf{Rationale Format} & \textbf{\texttt{edit\_acc}} & \textbf{\texttt{final\_acc}} & \textbf{\texttt{time\_taken(s)}} & \textbf{\texttt{edit\_acc}} & \textbf{\texttt{final\_acc}} & \textbf{\texttt{time\_taken(s)}} \\
    \midrule
    \texttt{markup\_mask} & 50.00 & 29.69$^{*|*|*|*}$ & 397.38 & 71.03 & 14.95$^{*|*|o|o}$ & 379.07 \\
    \texttt{annotated\_report} & 51.67 & 38.33$^{*|o|*|o}$ & 381.52 & 62.96 & \textbf{20.37}$^{*|*|*|*}$ & 426.04 \\
    \texttt{procedural} & 57.89 & \textbf{38.60}$^{*|o|o|*}$ & 359.75 & 55.77 & 8.65$^{*|*|*|*}$ & 434.24 \\
    \texttt{subquestions} & 49.21 & 36.51$^{*|*|*|o}$ & 375.45 & 59.05 & 14.29$^{o|*|*|*}$ & 425.14 \\
    \texttt{decision\_tree} & 56.25 & 37.50$^{*|o|*|o}$ & 397.56 & 69.52 & 17.14$^{o|*|*|*}$ & 494.42 \\
    \bottomrule
\end{tabular}
\caption{QA accuracy after patching generated rationales with human feedback and regenerating answers. We show here the \texttt{edit\_acc}, which is the percentage of examples for which the revised rationale successfully incorporates feedback and \texttt{final\_acc}, which measures the final answer accuracy after regeneration with the revised rationale. Statistical significance at $p<0.1$ is specified with $*$ (and $o$ if not significant) with paired bootstrap tests in the order of the remaining rows in the table.}
\label{tab:feedback_res}
\end{table*}

\subsection{Evaluation}

We first evaluate the effectiveness of feedback through edit accuracy (\texttt{edit\_acc}), where we manually label each revised rationale for whether it incorporates the feedback. Next, we compute final answer accuracy (\texttt{final\_acc}), where we check whether the final answer using the revised rationale is correct. We exclude all instances where the answer was leaked in feedback.


\subsection{Results}

We first show the results on standard decomposed QA for all rationale formats as well as standard answer generation (without rationales) on both Quoref and PubMedQA in Table~\ref{tab:initial_res}. These results show that decomposed QA models can be competitive with end-to-end models. Although they slightly underperform standard answer generation for Quoref, decomposed QA is better performing on PubMedQA. This suggests that decomposed QA is a promising modeling approach, while being predisposed to provide more faithful rationales.

Annotator labels of sufficiency and faithfulness (presented in Figure~\ref{fig:study1_both}) indicate that \texttt{annotated\_report} and \texttt{subquestions} have rationales that are most often sufficient for both datasets, while \texttt{markup\_mask} tends to lack most with sufficiency. On the other hand, extractive rationales from \texttt{markup\_mask} are labeled most faithful for Quoref (58\%), while \texttt{annotated\_report} is relatively faithful for both datasets.

Our main results for repairing rationales through feedback are in Table~\ref{tab:feedback_res}.
For Quoref, we find that rationale formats that expose the reasoning of the model are easier to repair through feedback. Interestingly, stricter formats with well-defined operations (such as \texttt{procedural}) are fairly effective for Quoref. On the other hand, for PubMedQA, rationale formats with more free-text components (such as \texttt{annotated\_report}) that can allow more flexible edits are most effective. This is likely because comprehending medical articles and making inferences based on them can involve nuances that are harder to express with strict rationale formats. For example, feedback written for one PubMedQA example (Table~\ref{tab:feedback_examples_pubmedqa}) mentioned that the rationale didn't consider the fact that the study did not consider a control group for testing their hypothesis. This is easily incorporated into the free-text nature of the \texttt{annotated\_report}, but is harder to incorporate in the \texttt{procedural} format.

We also note that when feedback is used to revise rationales that contain attributions, rationales can sometimes misquote sentences from the passage by hallucinating information that does not exist in the context. Although these revisions may result in correct answers, the rationales would be unfaithful, potentially decreasing user trust in the model.

In terms of annotator ease of providing feedback, we find that \texttt{markup\_mask} is easiest to provide feedback for, because it may be easy to verbalize when information is missing from the rationale. However, these judgements do not correlate with actual effectiveness of the feedback for rationale repair. This suggests annotator ease of providing feedback may not correlate with actual effectiveness of feedback.

\definecolor{bluez}{RGB}{210, 238, 255}
\begin{table*}[ht!]
\small
\scalebox{.8}{
\begin{tabular*}{19.6cm}{>{\raggedright\arraybackslash}p{3.5cm}>{\raggedright\arraybackslash}p{3.5cm}>{\raggedright\arraybackslash}p{3.5cm}>{\raggedright\arraybackslash}p{3.5cm}>{\raggedright\arraybackslash}p{3.5cm}}
\rowcolor{gray!15}
\multicolumn{5}{c}{\textbf{Q: What is the first name of the person whose son was a was a bachelor diplomat?}} \\
\rowcolor{gray!15} &&&& \\
\rowcolor{gray!15}
\textbf{Markup-and-mask} & \textbf{Annotated Report} & \textbf{Procedural} & \textbf{Subquestions} & \textbf{Decision Tree} \\ \toprule
\rowcolor{bluez!50}
\textbf{Error location}: Step 1 \newline \textbf{Issue}: The information is insufficient to answer the question and the inference drawn from the context is incorrect. \newline \textbf{Description}: Charles is the name of the son, and the question asks about the first name of the person who is the son's parent. \newline \textbf{Suggestion}: The rationale needs to find the name of the son and then look for the name of the son's parent in the preceding context. 
& \textbf{Error location}: Step 1 \newline \textbf{Issue}: The information is insufficient and the inference drawn from the context is incorrect. \newline \textbf{Description}: The quote and annotation in step 1 reveal who the son is, whereas the question is asking about the first name of the parent, not the son. \newline \textbf{Suggestion}: The rationale needs to find out who the son's parent is before providing their first name. 
& \textbf{Error location}: Step 1 \newline \textbf{Issue}: Insufficient information. \newline \textbf{Description}: Charles is the son who was a bachelor diplomat, and the question asks about the first name of Charles' parent. \newline \textbf{Suggestion}: The rationale needs to locate who Charles' parent is in the text and then provide their first name. 
& \textbf{Error location}: Q1 \newline \textbf{Issue}: The inference drawn from the context is incorrect. \newline \textbf{Description}: The question asks about the first name of the parent mentioned in the passage as having a son who was a bachelor diplomat is not Charles Spencer Cowper, and not the first name of the bachelor diplomat himself. \newline \textbf{Suggestion}: The rationale needs to find the bachelor diplomat's parent by looking at the context from the preceding sentences and then provide the parent's first name.
& \textbf{Error location}: Q2-Yes \newline \textbf{Issue}: The inference drawn from the context is incorrect. \newline \textbf{Description}: Charles Spencer Cowper is the bachelor diplomat, and the question asks about the first name of the person Charles' parent. \newline \textbf{Suggestion}: The rationale needs to look at previous sentences for the context of whose son Charles Spencer Cowper is and then provide that parent's first name. 
\end{tabular*}
}
\caption{Examples of feedback collected for different rationale formats for Quoref examples in study 1 (\S\ref{sec:study1}).}
\label{tab:feedback_examples_quoref}
\end{table*}

\definecolor{bluez}{RGB}{210, 238, 255}
\begin{table*}[ht!]
\small
\scalebox{.8}{
\begin{tabular*}{19.6cm}{>{\raggedright\arraybackslash}p{3.5cm}>{\raggedright\arraybackslash}p{3.5cm}>{\raggedright\arraybackslash}p{3.5cm}>{\raggedright\arraybackslash}p{3.5cm}>{\raggedright\arraybackslash}p{3.5cm}}
\rowcolor{gray!15}
\multicolumn{5}{c}{\textbf{Q: Is vitamin D insufficiency or deficiency related to the development of osteochondritis dissecans?}} \\
\rowcolor{gray!15} &&&& \\
\rowcolor{gray!15}
\textbf{Markup-and-mask} & \textbf{Annotated Report} & \textbf{Procedural} & \textbf{Subquestions} & \textbf{Decision Tree} \\ \toprule
\rowcolor{bluez!50}
\textbf{Error location}: Step 1 \newline \textbf{Issue}: Insufficient information \newline \textbf{Description}: There is not enough information to know whether vitamin D deficiency is related to the development of OCD lesions. Vitamin D could just be deficient in this population, and thus there could be many people with vitamin D deficiencies who never develop OCD lesions. \newline \textbf{Suggestion}: The rationale needs to consider the presence of a control group. This could be vitamin D levels before developing an OCD lesion and/or vitamin D levels from a group of people who never developed OCD lesions.
& \textbf{Error location}: Step 3 \newline \textbf{Issue}: Incorrect inferences drawn from Context \newline \textbf{Description}: Just because vitamin D levels are depleted amongst a group of OCD lesion patients does not mean that low vitamin D plays a role in the development of those lesions. For example, ... \newline \textbf{Suggestion}: The rationale needs to consider the presence of a control group. This could be a measurement of vitamin D levels before, during, and after developing OCD lesions.
& \textbf{Error location}: Step 1 \newline \textbf{Issue}: Insufficient information \newline \textbf{Description}: There is not a control group to compare the OCD patients' vitamin D levels to. Without a control group, we cannot know if Vitamin D is related to the development of OCD lesions. \newline \textbf{Suggestion}: The rationale needs to consider the presence of a control group. Whether the researchers measured Vitamin D levels and OCD prevalence in the general population.
& \textbf{Error location}: Q4 \newline \textbf{Issue}: Insufficient information \newline \textbf{Description}: The rationale says that the results suggest that low Vitamin D plays a role in the development of OCD lesions because vitamin D levels were depressed in a majority of the patients with OCD lesions. However, we do not have a control group/measurements and so cannot infer causality. \newline \textbf{Suggestion}: The rationale needs to consider the presence of a control group. This could be pre-OCD lesion Vitamin D levels in the same set of subjects.
& \textbf{Error location}: Q3-Yes \newline \textbf{Issue}: Incorrect inferences drawn from the context \newline \textbf{Description}: The model must have made an incorrect inference which caused them to not take the correct route down the decision tree and thus arrive at an incorrect answer. \newline \textbf{Suggestion}: Considering whether a control group was included would have allowed us to better understand any causality between vitamin D levels and developing OCD lesions.
\end{tabular*}
}
\caption{Examples of feedback collected for different rationale formats for PubMedQA examples in study 1 (\S\ref{sec:study1}).}
\label{tab:feedback_examples_pubmedqa}
\end{table*}

\section{Study 2: Evaluating User Perception of Rationales}
\label{sec:study2}

\subsection{Setup}

In the next study, we measure the extent to which different rationale formats enable users to \textit{understand} and \textit{trust} model responses. We sample examples where all 5 rationale formats have corresponding \textit{correct answers} for both datasets. 40 annotators completed this study for Quoref examples while 44 completed it for PubMedQA examples.

\subsection{Task}

We collect Likert ratings of interpretability and trustworthiness of rationales. In addition, we collect scalar judgements of the importance of different characteristics of rationales for annotators. Descriptions of Likert-scale options are in Figure~\ref{fig:interface2}.

\paragraph{Interpretability.} A rationale should facilitate in making the model's reasoning more transparent to an end user. This is measured by asking annotators how beneficial the rationale is in helping them understand the reasoning process followed by the model. It is elicited on a scale of (\textit{Very beneficial}, \textit{A bit beneficial}, \textit{Not beneficial at all}).

\paragraph{Trustworthiness.} In addition to improving user understanding, a rationale that makes a model's decision-making transparent should do so in a way that helps users trust model responses.
We ask annotators how likely they are to trust the model’s answer, if the rationale was provided along with the answer. The rating is elicited on a scale of (\textit{Very likely}, \textit{A bit likely}, \textit{A bit unlikely}, \textit{Not likely at all}).

\paragraph{Scalar judgements.} Next, we directly ask annotators for characteristics they value in rationales. They rate the following predefined rationale properties on a scale of 1-5 based on their importance:

\begin{itemize}
    \setlength{\itemsep}{0pt}
    \item \textit{Attribution}: Includes quotes from the context.
    \item \textit{Depth of reasoning}: Provides detailed insight into the reasoning process.
    \item \textit{Sequential reasoning}: Organized in a step-by-step manner.
    \item \textit{Strictness}: Contains well-defined steps, with strict input and output formats.
    \item \textit{Conciseness}: Brief and to the point.
\end{itemize}

\begin{figure*}[t!]
    \centering
    \includegraphics[width=2\columnwidth,height=7cm,keepaspectratio]{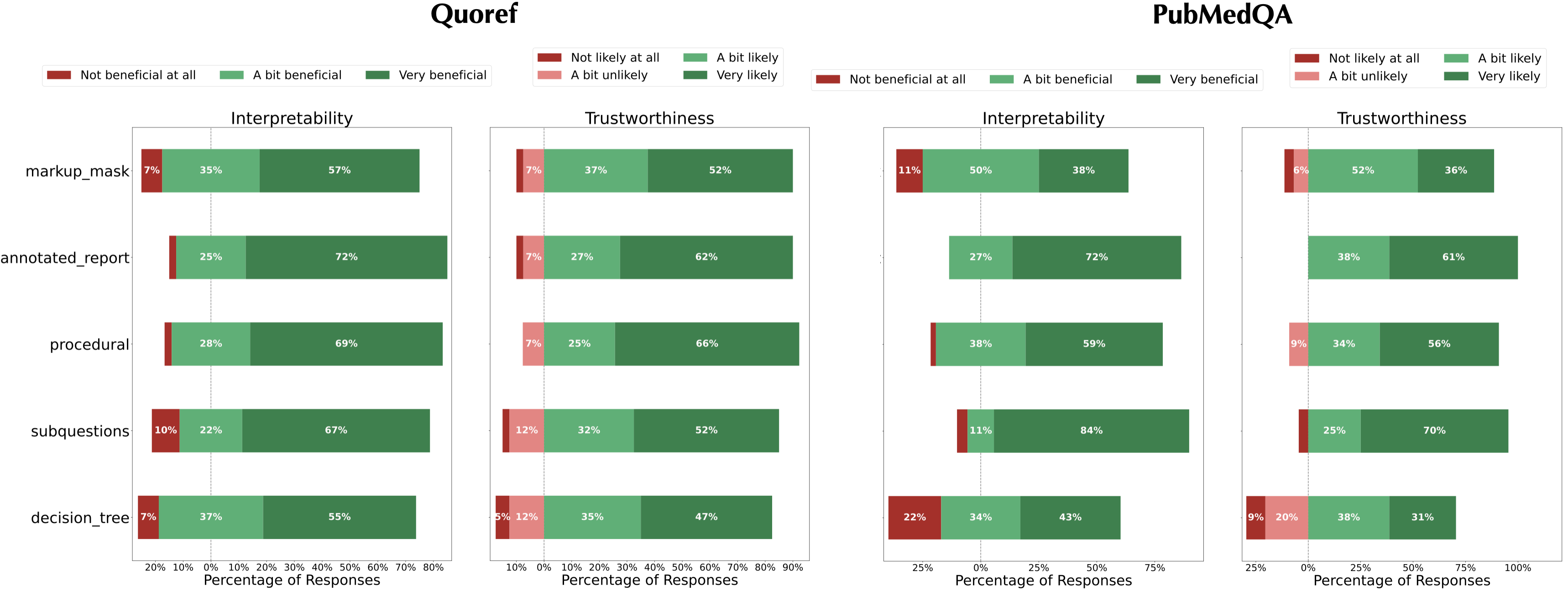}
    \caption{Likert distribution of the annotator judgements of interpretability \& trustworthiness for different rationale formats corresponding to correct answers (\S\ref{sec:study2}).}
    \label{fig:study2_both}
\end{figure*}

\subsection{Results}

Figure~\ref{fig:study2_both} shows the Likert distributions of judgements of interpretability and trustworthiness for all formats on both datasets. These suggest that rationales with attributions and a sufficient amount of depth (\texttt{annotated\_report} and \texttt{procedural}) are most easy to understand and trust for Quoref. On the other hand, \texttt{annotated\_report} and \texttt{subquestions} rate highest on both axes for PubMedQA. Our interpretation of these judgements is that to be easily understandable and trustworthy for users, rationales should provide sufficient insight into the model's reasoning process and be accompanied with attributions.

\begin{figure}[t!]
    \centering
    \includegraphics[width=\columnwidth,height=10cm,keepaspectratio]{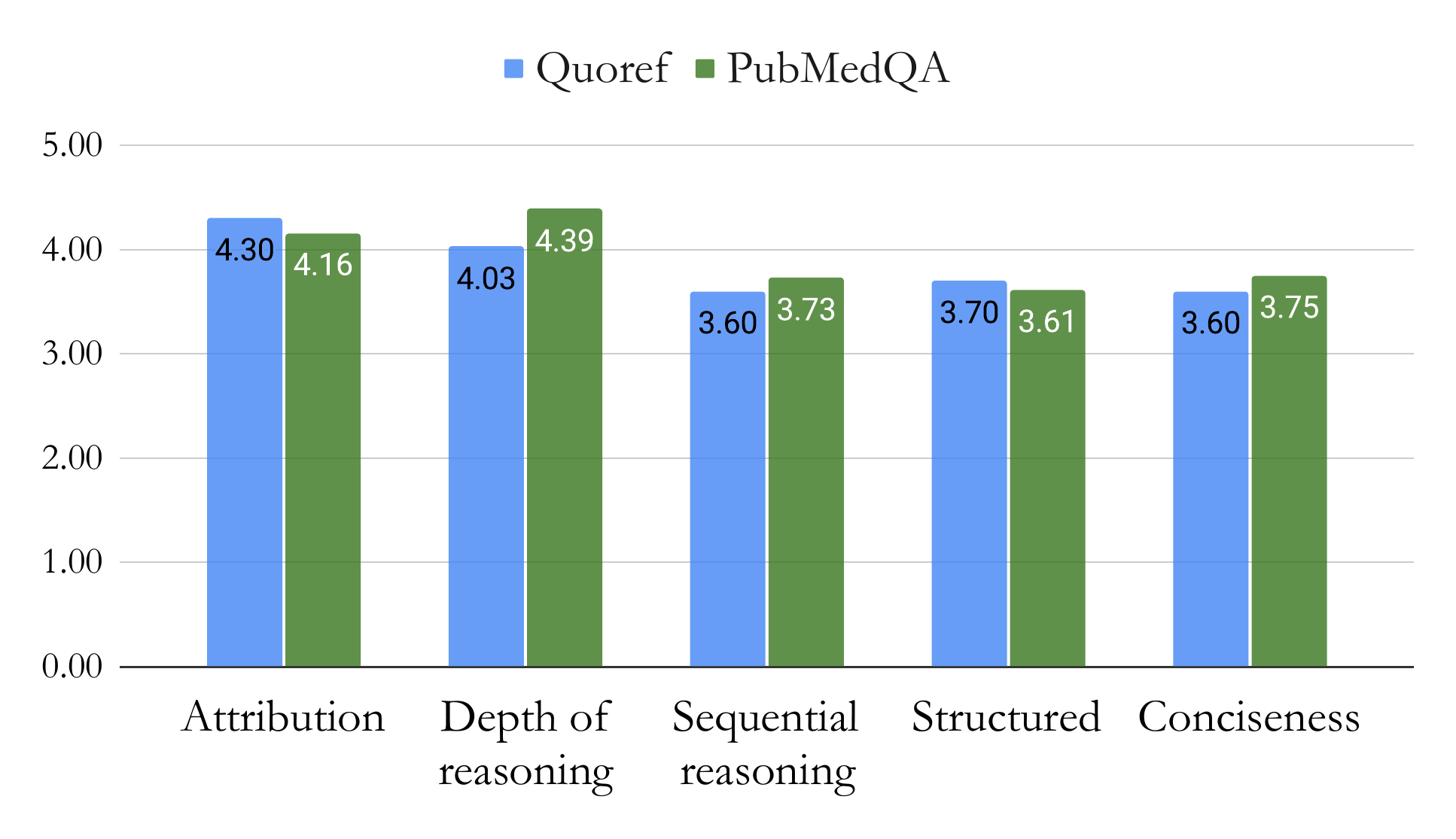}
    \caption{Scalar judgements of characteristics that annotators value in intermediate rationales (scale of 1-5).}
    \label{fig:scalar}
\end{figure}

Among the rationale properties presented to annotators, we find that attribution and depth of reasoning are rated as the most important properties of rationales. Figure~\ref{fig:scalar} shows averaged scalar judgements for different rationale properties. A clear conclusion from these judgements is that providing attributions in the form of extracted quotes to the context is essential to users. This is likely because the attributions ground the model's reasoning into the context. In addition, depth of reasoning is highly valuable to users, especially for PubMedQA questions, where they may value a logical and coherent description of the model's reasoning.

\section{Related Work}

\paragraph{Decomposed QA.} Although rationales from NLP models can be beneficial for users, there is recent evidence that shows that they are not always faithful to model responses \cite{ye2022unreliability, lyu2022towards, turpin2023language, lanham2023measuring}. Decomposed question answering systems break down the QA problem into two stages, that of generating an intermediate rationale, and then using only that rationale to generate the answer \cite{lei-etal-2016-rationalizing, eisenstein2022honest,radhakrishnan2023question}. This provides a stronger inductive bias to the model to be faithful to the rationale. Similar ideas have been pursued in other tasks such as object recognition \cite{koh2020concept}, image classification and text classification \cite{yeh2020completeness}. The precise format of the intermediate rationale that is optimal for human critiquability and interpretability is understudied. Our study is dedicated towards investigating the structure of this intermediate rationale.

\paragraph{Human feedback in NLP.} Providing human feedback to NLP models has proven to be an effective way to repair models and fix model behaviors \cite{fernandes2023bridging}. Feedback can allow users to convey example-level critiques about model predictions, which, when incorporated into models, encourage them to perform better. Prior work has explored using human feedback for improving text summarization \cite{stiennon2020learning, liu2022improving, scheurer2023training}, question answering \cite{gao2022simulating, li2022using}, semantic parsing \cite{iyer2017learning, elgohary2020speak, elgohary2021nl}, dialog generation \cite{shi2022life, ouyang2022training,xu2023learning}, machine translation \cite{kreutzer2018can} and image captioning \cite{fidler2017teaching}. Our work builds upon this prior work and investigates the effectiveness of human feedback for rationales provided by QA systems.

\paragraph{Rationales and explanations for NLP models.} There is a large body of prior work studying explanations to supplement outputs from NLP models, both for improving models \cite{hancock-etal-2018-training, lampinen2022can, wang2022rationale, zelikman2022star, zhou2023flame} and explaining model outputs to end users. Prior work has found that explanations can be beneficial to end users for understanding model responses \cite{deyoung-etal-2020-eraser, narang2020wt5, wiegreffe-etal-2022-reframing} as well as debugging models \cite{lertvittayakumjorn2021explanation,lamm-etal-2021-qed}. Prior work has also studied the impact of explanations on user trust \cite{papenmeier2022s} and performance of human-AI teams \cite{bansal2021does}. \citet{boyd-graber-etal-2022-human, jacovi-goldberg-2020-towards} provide useful guidelines to conduct human-centered and faithful evaluations of these explanations. We conduct another such evaluation that is centered on the format of model rationales presented to end users.

\section{Conclusion}

Our work analyzed how model-generated explanations or rationales should be formatted to be most amenable to repair through user feedback. We also collected qualitative judgements of how different formats enable users to understand and trust model outputs. We found that rationale formats significantly affect what rationales are amenable to be repaired through feedback. In terms of user perception of rationales, we find that some rationale formats, such as the \textit{annotated report}, are more favorable for enabling users to understand and trust model responses.
Finally, we find that among a few properties considered, attribution and depth of reasoning are the most important characteristics of rationales to users. We hope that this work can help researchers and practitioners alike make informed decisions about how to present language model responses and collect feedback from end users.

\section{Limitations}

\paragraph{Rationale Formats.} The rationale formats we consider are by no means exhaustive and there could be numerous other plausible formats for intermediate rationales. We choose a set of rationales that vary according to some well-defined properties (mentioned in Table~\ref{tab:formats}), that can allow us to form conclusions about the importance of those properties.

\paragraph{Feedback Structure.} We choose a feedback structure that encompasses a few crucial aspects of feedback highlighted in previous work. However, it may be possible that there are other types of feedback that show different trends in effectiveness across rationale formats.

\paragraph{Scope of QA problems.} We choose reading comprehension datasets where questions are formulated based on a given context. While these may not be representative of all forms of QA problems, we hope our findings can broadly inform practitioners about ways to present QA system responses to users (for instance, when deploying retrieval-augmented QA systems).

\section*{Acknowledgements}

First, we would like to thank the annotators who participated in our studies. We would also like to thank Elizabeth Sieber, Jacob Eisenstein, Weiqiu You and Eric Wong for helpful comments and discussions.

\bibliography{anthology,custom}
\bibliographystyle{acl_natbib}

\appendix

\clearpage
\section{Experimental Details}
\label{app:exps}
\begin{table*}[!ht]
\centering
\small
\begin{tabular}{lcccccc}
    \toprule
    & \multicolumn{3}{c}{Quoref} & \multicolumn{3}{c}{PubMedQA} \\
    \cmidrule(lr){2-4} \cmidrule(lr){5-7}
    \textbf{Rationale Format} & \textbf{\texttt{edit\_acc}} & \textbf{\texttt{final\_acc}} & \textbf{\texttt{time\_taken}} & \textbf{\texttt{edit\_acc}} & \textbf{\texttt{final\_acc}} & \textbf{\texttt{time\_taken}} \\
    \midrule
    \texttt{markup\_mask} & 70.37 & 62.96 & 340.29 & 64.58 & 33.33 & 389.79 \\
    \texttt{annotated\_report} & 61.29 & 58.06 & 290.17 & 58.06 & \textbf{45.16} & 447.90 \\
    \texttt{procedural}  & 52.94 & 58.82 & 345.55 & 51.28 & 33.33 & 333.80 \\
    \texttt{subquestions}  & 81.81 & \textbf{72.72} & 316.71 & 65.12 & 30.23 & 348.57 \\
    \texttt{decision\_tree}  & 66.67 & 38.10 & 340.39 & 88.64 & 13.64 & 465.39 \\
    \bottomrule
\end{tabular}
\caption{Results after patching generated rationales with human feedback for examples where the answer is wrong for 3 rationale formats. We show here the \texttt{edit\_acc}, which measures if the revised rationale successfully incorporates feedback and \texttt{final\_acc}, which measures the final accuracy after regeneration with the revised rationale.}
\label{tab:feedback_res_3wrong}
\end{table*}

\begin{table*}[ht!]
    \centering
    \begin{tabular}{p{2\columnwidth}}
        \toprule
        \textbf{X2R Prompt (\texttt{markup\_mask})} \\
        \midrule
        \raggedright \texttt{Extract the most relevant 1-2 sentences from the context as a rationale sufficient to answer the question. Also resolve any ambiguous terms and coreferences in the extracted sentences to make them standalone. The relevant sentences should be sufficient to determine the answer to the question.\linebreak \linebreak Context: [CONTEXT] \linebreak \linebreak Question: [QUESTION] \linebreak \linebreak Rationale: \linebreak}
        \hrule
    \end{tabular}
    \caption{$X2R$ prompt for the \texttt{markup\_mask} format.}
    \label{tab:x2r_prompt_markup_mask}
\end{table*}

\begin{table*}[ht!]
    \centering
    \begin{tabular}{p{2\columnwidth}}
        \toprule
        \textbf{X2R Prompt (\texttt{annotated\_report})} \\
        \midrule
        \raggedright \texttt{Generate a rationale that is helpful and sufficient to answer the question. The rationale should contain a list of extracted phrases from the context and the conclusion drawn from each phrase. Try to include no more than 5 extracted phrases. \linebreak \linebreak Context: [CONTEXT] \linebreak \linebreak Question: [QUESTION] \linebreak \linebreak Annotations: \linebreak}
        \hrule
    \end{tabular}
    \caption{$X2R$ prompt for the \texttt{annotated\_report} format.}
    \label{tab:x2r_prompt_annotated_report}
\end{table*}

\begin{table*}[ht!]
    \centering
    \begin{tabular}{p{2\columnwidth}}
        \toprule
        \textbf{X2R Prompt (\texttt{procedural})} \\
        \midrule
        \raggedright \texttt{Construct a structured Plan for answering the Question, that should provide a sequential process for finding the answer. 
The Plan should not directly answer the Question but only provide the reasoning.
You can use the following operations in the plan:\linebreak
- Extract-relevant-sentences: Extract relevant sentences from the passage that are sufficient to answer the question. The extracted sentences should include the necessary information to answer the question accurately.\linebreak
- Disambiguate-question-entity(s): Determine the specific entity or phrase that the string s in the question refers to. Clarify any ambiguous terms or references to ensure a precise understanding.\linebreak
- Disambiguate-plan-entity(s): Identify the entity or phrase that the string s in the plan refers to. Resolve any ambiguity within the plan by specifying the relevant entities explicitly.\linebreak
- Locate-entity(s): Generate a subquestion s that is important to answer the original question without simply repeating the original question. Determine the exact entity or phrase that provides the answer to the subquestion s.\linebreak \linebreak Context: [CONTEXT] \linebreak \linebreak Question: [QUESTION] \linebreak \linebreak Plan: \linebreak}
        \hrule
    \end{tabular}
    \caption{$X2R$ prompt for the \texttt{procedural} format.}
    \label{tab:x2r_prompt_procedural}
\end{table*}

\begin{table*}[ht!]
    \centering
    \begin{tabular}{p{2\columnwidth}}
        \toprule
        \textbf{X2R Prompt (\texttt{subquestions})} \\
        \midrule
        \raggedright \texttt{Form subquestions required to answer the given question based on the passage. You cannot repeat the given question as a subquestion. The formed subquestions and their answers should be sufficient to answer the given question. Try to form no more than 5 subquestions.\linebreak \linebreak Context: [CONTEXT] \linebreak \linebreak Question: [QUESTION] \linebreak \linebreak Subquestions: \linebreak}
        \hrule
    \end{tabular}
    \caption{$X2R$ prompt for the \texttt{subquestions} format.}
    \label{tab:x2r_prompt_subquestions}
\end{table*}

\begin{table*}[ht!]
    \centering
    \begin{tabular}{p{2\columnwidth}}
        \toprule
        \textbf{X2R Prompt (\texttt{decision\_tree})} \\
        \midrule
        \raggedright \texttt{Generate a decision tree-based rationale to answer the question. The decision tree should be sufficient to answer the question. However, it should not answer the question directly. Try to form no more than 5 subquestions.\linebreak \linebreak Context: [CONTEXT] \linebreak \linebreak Question: [QUESTION] \linebreak \linebreak Decision Tree: \linebreak}
        \hrule
    \end{tabular}
    \caption{$X2R$ prompt for the \texttt{decision\_tree} format.}
    \label{tab:x2r_prompt_decision_tree}
\end{table*}

\begin{table*}[ht!]
    \centering
    \begin{tabular}{p{2\columnwidth}}
        \toprule
        \textbf{R2Y Prompt (\texttt{markup\_mask})} \\
        \midrule
        \raggedright \texttt{Use these extracted relevant sentences from a passage to answer the question.\linebreak \linebreak Relevant sentences: [RATIONALE] \linebreak \linebreak Question: [QUESTION] \linebreak \linebreak Answer: \linebreak}
        \hrule
    \end{tabular}
    \caption{$R2Y$ prompt for the \texttt{markup\_mask} format.}
    \label{tab:r2y_prompt_markup_mask}
\end{table*}

\begin{table*}[ht!]
    \centering
    \begin{tabular}{p{2\columnwidth}}
        \toprule
        \textbf{R2Y Prompt (\texttt{annotated\_report})} \\
        \midrule
        \raggedright \texttt{You are given an annotated rationale from a passage as context. The annotations are in the format of a list of extracted phrases from the context and the conclusion drawn from each phrase. Answer the question based on the rationale alone.\linebreak \linebreak Rationale: [RATIONALE] \linebreak \linebreak Question: [QUESTION] \linebreak \linebreak Answer: \linebreak}
        \hrule
    \end{tabular}
    \caption{$R2Y$ prompt for the \texttt{annotated\_report} format.}
    \label{tab:r2y_prompt_annotated_report}
\end{table*}

\begin{table*}[ht!]
    \centering
    \begin{tabular}{p{2\columnwidth}}
        \toprule
        \textbf{R2Y Prompt (\texttt{procedural})} \\
        \midrule
        \raggedright \texttt{Answer the Question based on the Plan-based Rationale. The Plan gives a sequential process of finding the answer. The following operations can be used in a plan: <Skipped for brevity>.\linebreak \linebreak Plan: [RATIONALE] \linebreak \linebreak Question: [QUESTION] \linebreak \linebreak Answer: \linebreak}
        \hrule
    \end{tabular}
    \caption{$R2Y$ prompt for the \texttt{procedural} format.}
    \label{tab:r2y_prompt_procedural}
\end{table*}

\begin{table*}[ht!]
    \centering
    \begin{tabular}{p{2\columnwidth}}
        \toprule
        \textbf{R2Y Prompt (\texttt{subquestions})} \\
        \midrule
        \raggedright \texttt{Answer the given Question solely based on the Subquestions and their answers. The answer can always be found from the Subquestions so make your best guess.\linebreak \linebreak Subquestions: [RATIONALE] \linebreak \linebreak Question: [QUESTION] \linebreak \linebreak Answer: \linebreak}
        \hrule
    \end{tabular}
    \caption{$R2Y$ prompt for the \texttt{subquestions} format.}
    \label{tab:r2y_prompt_subquestions}
\end{table*}

\begin{table*}[ht!]
    \centering
    \begin{tabular}{p{2\columnwidth}}
        \toprule
        \textbf{R2Y Prompt (\texttt{decision\_tree})} \\
        \midrule
        \raggedright \texttt{Answer the Question solely based on the Decision Tree-based Rationale.\linebreak \linebreak Decision Tree: [RATIONALE] \linebreak \linebreak Question: [QUESTION] \linebreak \linebreak Answer: \linebreak}
        \hrule
    \end{tabular}
    \caption{$R2Y$ prompt for the \texttt{decision\_tree} format.}
    \label{tab:r2y_prompt_decision_tree}
\end{table*}

\paragraph{Prompts.} The prompts used to generate rationales for all formats are provided in Tables~\ref{tab:x2r_prompt_markup_mask}, \ref{tab:x2r_prompt_annotated_report}, \ref{tab:x2r_prompt_procedural}, \ref{tab:x2r_prompt_subquestions} and \ref{tab:x2r_prompt_decision_tree}, while the prompts used to generate answers are provided Tables~\ref{tab:r2y_prompt_markup_mask}, \ref{tab:r2y_prompt_annotated_report}, \ref{tab:r2y_prompt_procedural}, \ref{tab:r2y_prompt_subquestions} and \ref{tab:r2y_prompt_decision_tree}. For generating answers for PubMedQA, we modify the prompt same way as previous work \cite{lievin2022can}, by transforming it into a multiple-choice question. For revising rationales, we use a similar format as these prompts but also includes the following string in the instruction -- 

\texttt{Correct the given Rationale based on the Feedback. The Feedback first points out the Error Location, then mentions the Issue and gives a Description of the issue, and finally provides a Suggestion to correct the given Rationale. The Rationale is required to be sufficient to answer the Question on its own and faithful to the Context.}.

\paragraph{Hyperparameter settings.} Rationales and answers were sampled from \texttt{gpt-3.5-turbo} with a temperature of 0.0 and a maximum length of 512 tokens when sampling rationales, and 64 tokens when sampling answers. For prompting models, we sample few-shot exemplars using BM25 \cite{robertson2009probabilistic} from a set of 100 manually labeled examples with rationales. We sample as many exemplars as can fit within the maximum sequence length (4096) of the model. 

\section{Annotation Details}
\label{app:annotation}

\paragraph{Annotator backgrounds.} For both studies, annotators were recruited from Prolific\footnote{\url{www.prolific.co}}, and required to be fluent in English. They were required to have at least 100 accepted submissions and an approval rate of at least 99\%. They were also required to have at least a bachelor's degree.

Annotators for the Quoref task were based in UK, USA, Australia, Ireland, Canada or New Zealand. Annotators for the PubMedQA task were based in UK, USA, Ireland, Germany, France, Australia, Canada, Denmark, Netherlands, Switzerland, Norway, Portugal or Sweden. These annotators were additionally required to be employed in the healthcare/medicine sector.

\paragraph{Annotation costs.} In both studies, annotators were compensated at the rate of \$15 per hour with additional bonuses when annotators spent more time than we anticipated.

\begin{figure*}[t!]
        \begin{adjustwidth}{-1.5cm}{-1.5cm}
        \vspace{-1cm}
        \centering
        \scalebox{1.2}{\includegraphics[width=2\columnwidth]{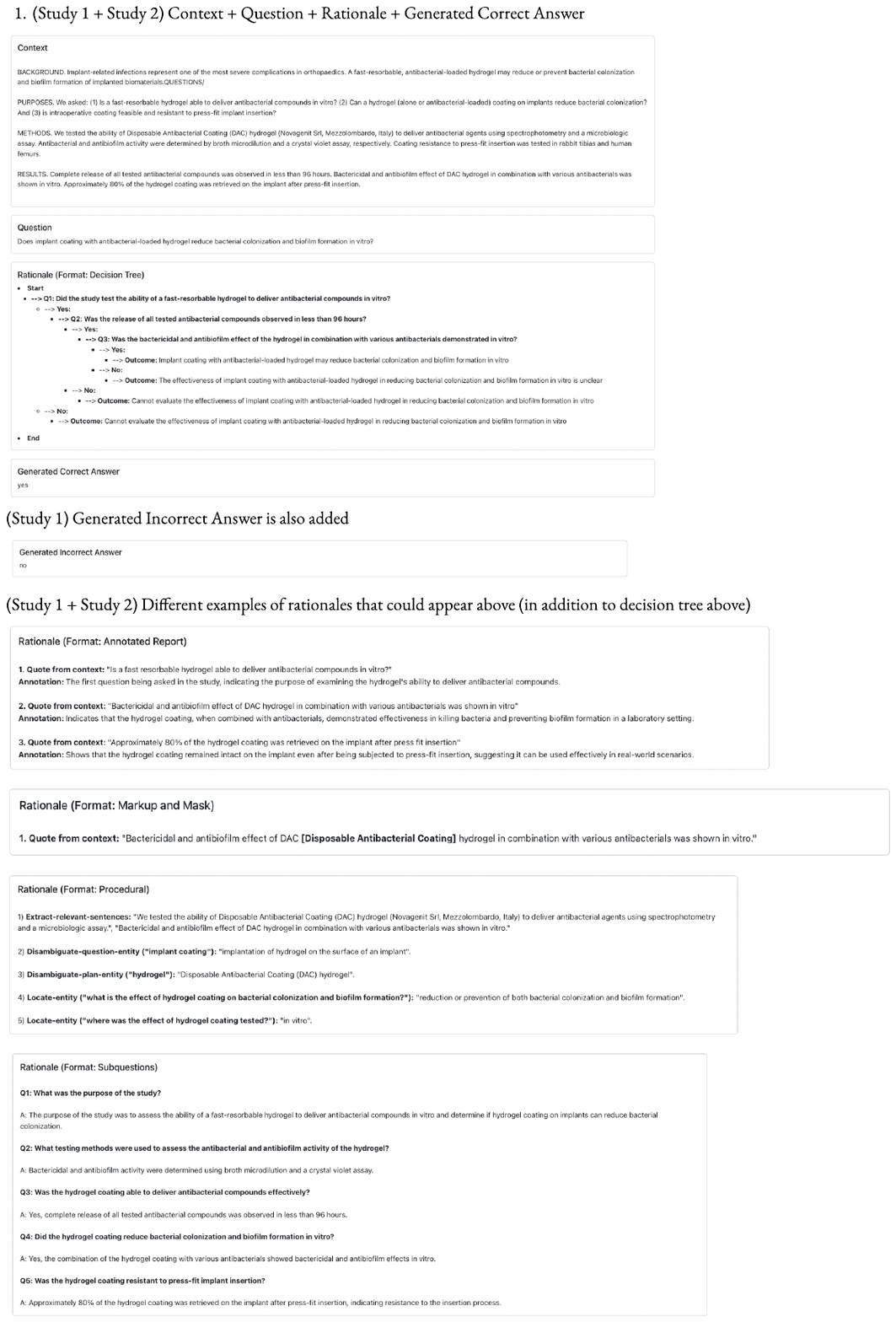}}
        \caption{Screenshots of the interface (1).}
        \label{fig:interface1}
    \end{adjustwidth}
\end{figure*}

\begin{figure*}[t!]
    \begin{adjustwidth}{-1.5cm}{-1.5cm}
        \vspace{-1cm}
        \centering
        \scalebox{1.2}{\includegraphics[width=2\columnwidth]{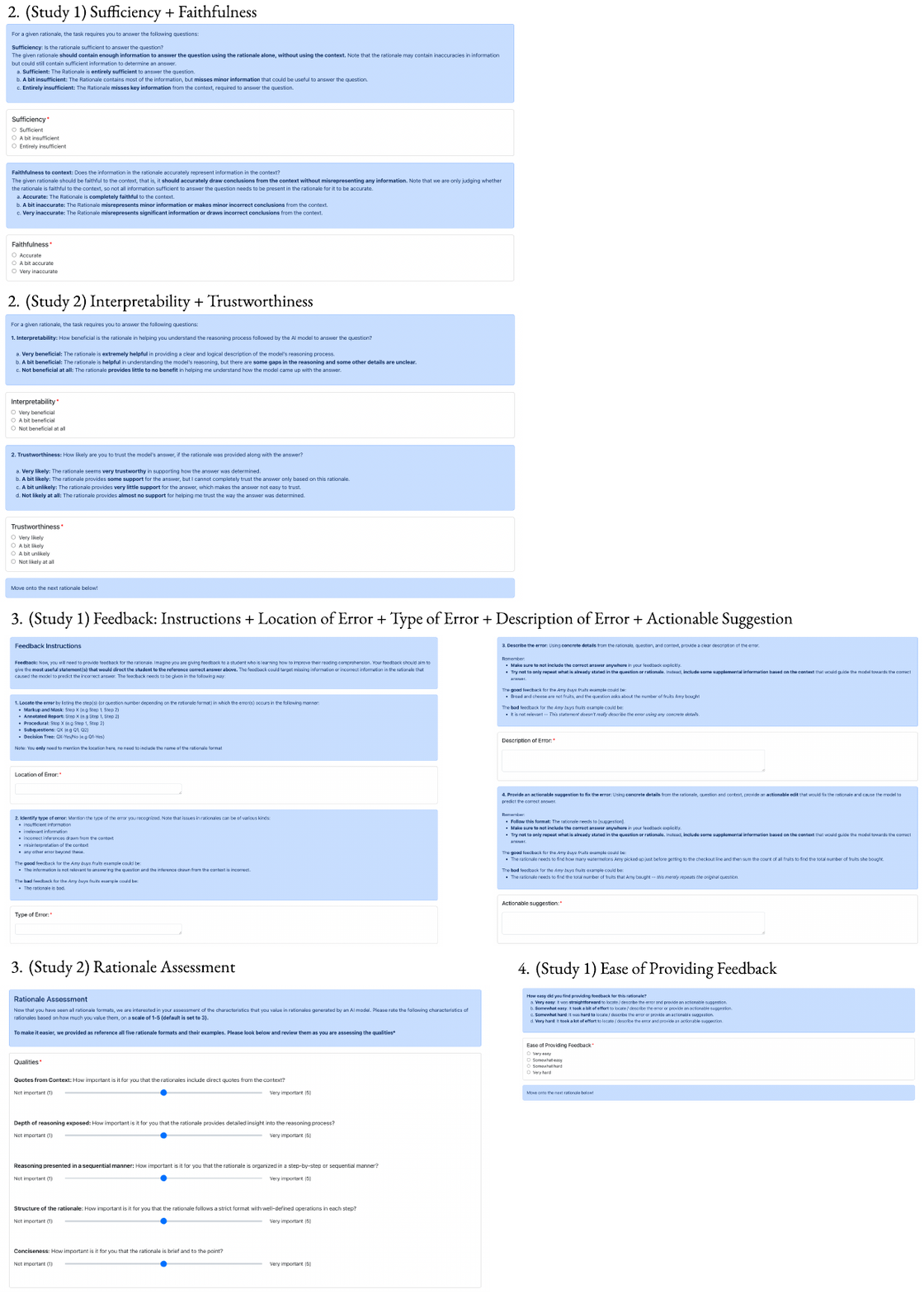}}
        \caption{Screenshots of the interface (2-3).}
        \label{fig:interface2}
    \end{adjustwidth}
\end{figure*}

\paragraph{Annotation interface.} Figures~\ref{fig:interface1} and \ref{fig:interface2} show screenshots of our annotation interface for both Study 1 and Study 2 in the order the task was presented to annotators.

\section{Additional Results}
\label{app:additional}

\begin{figure*}[t!]
    \centering
    \includegraphics[width=2\columnwidth]{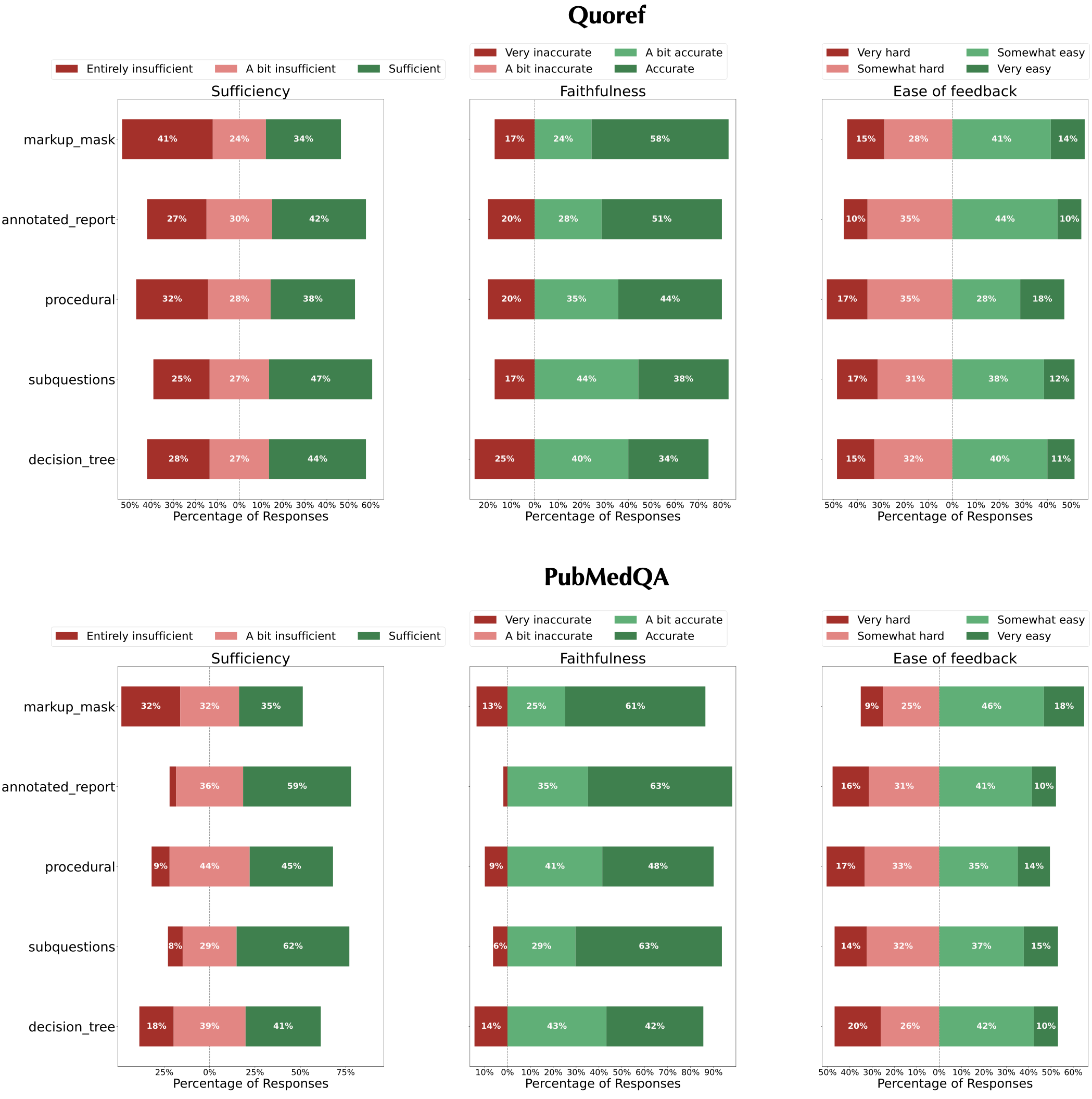}
    \caption{Likert distribution of the sufficiency \& faithfulness for different rationale formats, as well as ease of writing feedback. (\S\ref{sec:study1}).}
    \label{fig:study1_both}
\end{figure*}

Table~\ref{tab:feedback_res_3wrong} shows the effectiveness of feedback with examples where 3 rationale formats get the answer wrong.
Figure~\ref{fig:study1_both} shows the Likert distribution of sufficiency, faithfulness and ease of providing feedback for all rationale formats for both datasets.

\end{document}